\documentclass{article}
\usepackage{ijcai23}

\usepackage{times}
\usepackage{soul}
\usepackage{url}
\usepackage[hidelinks]{hyperref}
\usepackage[utf8]{inputenc}
\usepackage[small]{caption}
\usepackage{graphicx}
\usepackage{amsmath}
\usepackage{amsthm}
\usepackage{booktabs}
\usepackage{algorithm}
\usepackage{algorithmic}
\usepackage[switch]{lineno}

\usepackage{caption}
\usepackage{graphicx}
\usepackage{subfigure}
\usepackage{float}
\usepackage{enumerate}
\usepackage{empheq}
\usepackage{amssymb}
\usepackage{amsfonts}

\usepackage{tabularx}
\usepackage{booktabs}
\usepackage{makecell}
\usepackage{mathtools}
\usepackage{multirow}
\usepackage{mathtools}
\usepackage{bbm}

\usepackage{array}
\usepackage[table]{xcolor}
\usepackage{nicematrix}
\usepackage{subfigure}
\usepackage{lipsum}
\usepackage{dsfont}

\DeclareMathOperator*{\argmin}{argmin}
\DeclareMathOperator*{\argmax}{argmax}

\title{An Offline Time-aware Apprenticeship Learning Framework for \\ 
Evolving Reward Functions}

\author{
    Paper ID: \#3122
}

\author{
Xi Yang  \and
Ge Gao \and
Min Chi 
\affiliations
Department of Computer Science, North Carolina State University \\
\emails
\{yxi2, ggao5, mchi\}@ncsu.edu
}

\begin{document}

\maketitle

\begin{abstract}
Apprenticeship learning (AL) is a process of inducing effective decision-making policies via observing and imitating experts' demonstrations. Most existing AL approaches, however, are not designed to cope with the \emph{evolving reward functions} commonly found in human-centric tasks such as healthcare, where \emph{offline} learning is required. In this paper, we propose an \emph{offline} Time-aware Hierarchical EM Energy-based Sub-trajectory (THEMES) AL framework to tackle the evolving reward functions in such tasks. The effectiveness of THEMES is evaluated via a challenging task -- sepsis treatment. The experimental results demonstrate that THEMES can significantly outperform competitive state-of-the-art baselines. 

\end{abstract}

\section{Introduction}

Apprenticeship learning (AL) aims at inducing decision-making policies via observing and imitating experts' demonstrations \cite{abbeel2004apprenticeship}. It is particularly applicable for human-centric tasks such as healthcare, where assigning accurate \emph{reward functions} is extremely challenging \cite{abbeel2004apprenticeship,babes2011apprenticeship}. The existing AL approaches are commonly \emph{online}, requiring interacting with the environment iteratively for collecting new data and then updating the model accordingly \cite{abbeel2004apprenticeship,ziebart2008maximum,ho2016generative,finn2016guided}. However, in many real-world human-centric tasks, \textit{e.g.}, healthcare, the execution of a potential bad policy can be not only unethical but also illegal \cite{levine2020offline}. Therefore, it is highly desired to induce the policies purely based on the demonstrated behaviors in an \emph{offline} manner.

A recently proposed energy-based distribution matching (EDM) approach \cite{jarrett2020strictly} has advanced the state-of-the-art in \emph{offline} AL. When applying EDM to induce policies in real-world human-centric tasks, there are two major challenges:
1)  EDM assumes that all demonstrations are generated using a homogeneous policy with a \emph{single} reward function, whereas when humans make decisions, the reward function can evolve over time. As an example, when investing in retirement accounts, people tend to turn from aggressive to conservative investments as they age, resulting in a shift from taking risks to avoiding them. The same can be said in healthcare, where clinicians will adapt their interventions when treating patients in different stages during the disease progression, such as common infection versus severe shock \cite{wang2022hierarchical}. By considering the evolving reward functions, we propose a hierarchical AL that automatically partitions the trajectory into sub-trajectories (stages) to learn the decision-making policies at various progressive stages. 
2) EDM does not consider the irregular time intervals between consecutive states in the demonstrations. Measurements in human-centric tasks are commonly acquired with \emph{irregular intervals}. For example, the intervals in electronic healthcare records (EHRs) vary from seconds to days \cite{baytas2017patient}. When a patient is in a severe condition, states are likely to be recorded more frequently than when a patient is in a relatively ``healthier" condition. Hence such varying intervals can reveal patients' health status on certain impending conditions, and it is important to consider the time intervals between temporal states to capture latent disease progressive patterns. To address this issue, we incorporate \emph{time-awareness} into the AL framework, for better capturing the latent progressive patterns to induce more accurate policies.

We propose a \textbf{\emph{\underline{T}}}ime-aware \textbf{\emph{\underline{H}}}ierarchical \textbf{\emph{\emph{\underline{EM}}}} \textbf{\emph{\underline{E}}}nergy-based \textbf{\emph{\underline{S}}}ubtrajectory (THEMES) AL framework to handle the \emph{multiple} reward functions \emph{evolving} over time in an \emph{offline} manner. THEMES consists of two main components that work together iteratively: 
1) a Reward-regulated \underline{M}ulti-series \underline{T}ime-aware \underline{T}oeplitz \underline{I}nverse \underline{C}ovariance-based \underline{C}lustering (RMT-TICC), which incorporates the \emph{time-awareness} for automatically partitioning the trajectories and 2) an \emph{offline} \underline{EM} \underline{E}nergy-based \underline{D}istribution \underline{M}atching (EM-EDM) to cluster the partitioned sub-trajectories and induce policy for each cluster. To tackle the multiple reward functions \emph{evolving} over time, the key challenge lies in how to effectively partition the trajectories. By incorporating a \textit{reward regulator} which characterizes the decision-making patterns learned by EM-EDM, more accurate sub-trajectory partitions can be achieved by RMT-TICC; Meanwhile, accurate partitions derived by RMT-TICC can, in turn, enhance the effectiveness of EM-EDM to induce more accurate policies. THEMES is compared against five competitive state-of-the-art baselines on an extremely challenging task -- sepsis treatment. 

Sepsis is a life-threatening organ dysfunction and a leading cause of death worldwide. Septic shock, the most severe complication of sepsis, leads to a mortality rate as high as 50\% \cite{martin2003epidemiology}. Contrarily, 80\% of sepsis deaths can be prevented with timely interventions \cite{kumar2006duration}. Sepsis generally involves various progressive stages, and clinicians are supposed to tailor their interventions in different stages \cite{singer2016third}. However, it is usually challenging to identify different stages due to the complex pathological process, since each stage involves more refined stages, and different stages may co-occur \cite{stearns2011pathogenesis}. Our proposed THEMES can discover such stages automatically and induce different intervention policies accordingly. The major contributions of this work are three-folded:

\noindent $\bullet$ By incorporating the \emph{time-awareness} and \emph{decision-making patterns} in RMT-TICC, the fine-grained sub-trajectories can be derived to indicate different progressive stages.

\noindent $\bullet$ By utilizing our offline EM-EDM, decision-making policies over different progressive stages can be efficiently induced, without the need of rolling out the policies iteratively.

\noindent $\bullet$ THEMES shows capability in modeling complex human-centric decision-making tasks with \emph{multiple} reward functions \emph{evolving} over time. The significant performance in healthcare sheds some light on assessing the clinicians' interventions and assisting them with timely personalized treatments.

\section{Related Work}

\subsection{Offline AL}

As a classic offline AL approach, \emph{behavior cloning} \cite{syed2012imitation,raza2012teaching} learns a mapping from states to actions by greedily imitating demonstrated experts' optimal behaviors \cite{pomerleau1991efficient}. To better capture the data distribution in experts' demonstrations, some \textit{inverse reinforcement learning (IRL)-based} \cite{abbeel2004apprenticeship,ziebart2008maximum} as well as \textit{adversarial imitation learning-based} \cite{ho2016generative} methods have been proposed. 

\emph{IRL-based} methods generally involve an iterative loop for 1) inferring a reward function, 2) inducing a policy by traditional RL, 3) rolling out the learned policy, and then 4) updating the reward parameters based on the divergence between the roll-out behaviors and experts' demonstrations. When re-purposing the IRL methods that follow the above online manner to an offline setting, their original computational and theoretical disadvantages will be inherited accordingly. Additionally, IRL-based methods usually model the reward via a certain tractable format, \textit{e.g.}, the linear function \cite{abbeel2004apprenticeship,ziebart2008maximum,babes2011apprenticeship}, mapping from states/state-action pairs to reward values. Besides, to avoid rolling out the learned policy, some batch-IRL have been proposed \cite{raghu2017continuous}, while they usually require off-policy evaluation to update the reward function, which is itself nontrivial with imperfect solutions. 

\emph{Adversarial imitation learning-based} methods typically require iteratively learning a generator to roll out the policy and a discriminator to distinguish the learned behaviors from the experts' demonstrations. Under the batch setting, to avoid rolling out the policy, some off-policy learning methods have been proposed based on off-policy actor-critic \cite{kostrikov2018discriminator,kostrikov2019imitation}. However, these methods inherit the complex alternating max-min optimization from general adversarial imitation learning \cite{ho2016generative}. 

Recently, Jarrett \textit{et al.} proposed and evaluated an \emph{EDM} \cite{jarrett2020strictly} on a wide range of benchmarks, \textit{e.g.}, Acrobot, LunarLander, and BeamRider \cite{brockman2016openai}, and they show  EDM can better model the offline AL compared to \emph{IRL-based} and \emph{adversarial imitation learning-based} methods. In this work, EDM serves as one of our baselines.

\subsection{AL with Multiple Reward Functions}

Several AL methods have been proposed to handle multiple reward functions which assumes the \emph{reward function to be different across trajectories, while for each trajectory, the reward function remains the same}. Dimitrakakis and Rothkopf proposed a Bayesian multi-task IRL, which models the heterogeneity of reward functions by formalizing it as a statistical preference elicitation, via a joint reward-policy prior \cite{dimitrakakis2011bayesian}. Choi and Kim integrated a Dirichlet process mixture model into Bayesian IRL to cluster the demonstrations \cite{choi2012nonparametric}. Using a Bayesian model, they incorporated the domain knowledge of multiple reward functions. Likewise, Arora \textit{et al.} combined the Dirichlet process with a maximum entropy IRL to get demonstration clusters with respective reward functions \cite{arora2021min}. Babes \textit{et al.} presented an EM-based IRL approach that clusters trajectories based on their different reward functions \cite{babes2011apprenticeship}. As a component of the EM-based IRL, a maximum-likelihood IRL uses gradient ascent to optimize the reward parameters, which has shown success in efficiently identifying unknown reward functions.

To deal with \emph{dynamic evolving} reward functions \emph{varying over time within and across trajectories}, some AL methods have been proposed but they require either \emph{online} interactions or off-policy evaluation, which may introduce more variance than one can afford \cite{jarrett2020strictly}. Krishnan \textit{et al.} developed a hierarchical inverse reinforcement learning (HIRL) \cite{krishnan2016hirl} for long-horizon tasks with delayed rewards. Each trajectory is assumed to complete a certain task and can be decomposed into sub-tasks with shorter horizons based on transitions that are consistent across demonstrations. By identifying the changes in local linearity, the sub-tasks can be identified, which will be further utilized for constructing the local rewards with respect to the global sequential structure.
Later, Hausman \textit{et al.} proposed a multi-modal imitation learning (MIL) framework to segment and imitate skills (\textit{i.e.}, policies) from unlabelled and unstructured demonstrations by conducting policy segmentation and imitation learning jointly \cite{hausman2017multi}. Based on the generative adversarial imitation learning \cite{ho2016generative}, they augmented the input via a latent intention with a categorical or uniform distribution to model the heterogeneity of policies varying over time.
Afterward, Wang \textit{et al.} proposed to learn sub-trajectories via contextual bandits \cite{wang2020multi}. They induce the policies over sub-trajectories by both behavior cloning and generative adversarial imitation learning, while the latter is similar to MIL, requiring either \emph{online} interactions or off-policy evaluations to update the model. More recently, Wang \textit{et al.} presented a hierarchical imitation learning model \cite{wang2022hierarchical} to jointly learn latent high-level policies and sub-policies for each individual sub-task from experts' demonstrations, where the sub-trajectories are assumed to have \emph{fixed length}. Additionally, \cite{nguyen2015inverse} and \cite{ashwood2022dynamic} presented two offline AL methods for involving reward functions. Different from our THEMES, both of them take \emph{discrete states} as input and approximate reward functions through \emph{linear functions}. Therefore, in this work, we take HIRL \cite{krishnan2016hirl} and MIL \cite{hausman2017multi} as baselines.

\subsection{Partitioning Trajectories}

Classic methods for partitioning trajectories can be categorized into \emph{distance-based} \cite{cuturi2011fast,li2012visualizing} and \emph{model-based} \cite{severson2020personalized,reynolds2009gaussian,smyth1997clustering,hallac2017toeplitz}. Recently, Severson \textit{et al.} employed a hidden Markov model-based method to learn the Parkinson's progression, which encoded prior knowledge to learn the latent states and then used post-hoc analysis to interpret the states \cite{severson2020personalized}. Hallac \textit{et al.} proposed a Toeplitz inverse covariance-based clustering \cite{hallac2017toeplitz}, which employs inverse covariance matrices and constrains these matrices to be block-wise Toeplitz to model the time-invariant structural patterns in each cluster. It outperformed both \emph{distance-based} methods, \textit{e.g.}, dynamic time warping \cite{cuturi2011fast} or rule-based motif discovery \cite{li2012visualizing}), and \emph{model-based} methods, \textit{e.g.}, Gaussian mixture models \cite{reynolds2009gaussian} or hidden Markov models on real-life applications, \textit{e.g.}, analyzing physical activities for patients with Alzheimer's \cite{li2018tatc} and segmenting the critical stages for sepsis patients \cite{gao2022reinforcement}. More recently, Yang \textit{et al.} proposed a multi-series time-aware Toeplitz inverse covariance-based clustering (MT-TICC) by extending the \cite{hallac2017toeplitz} to cope with multi-series inputs and irregular intervals. Given benchmarks, MT-TICC showed superior performance over both classic and state-of-the-art methods. In this work, we further extended MT-TICC by proposing a reward-regulated MT-TICC (RMT-TICC).

\section{Our Methodology: THEMES}

\begin{figure}
    \centering
    \includegraphics[width=6.5cm]{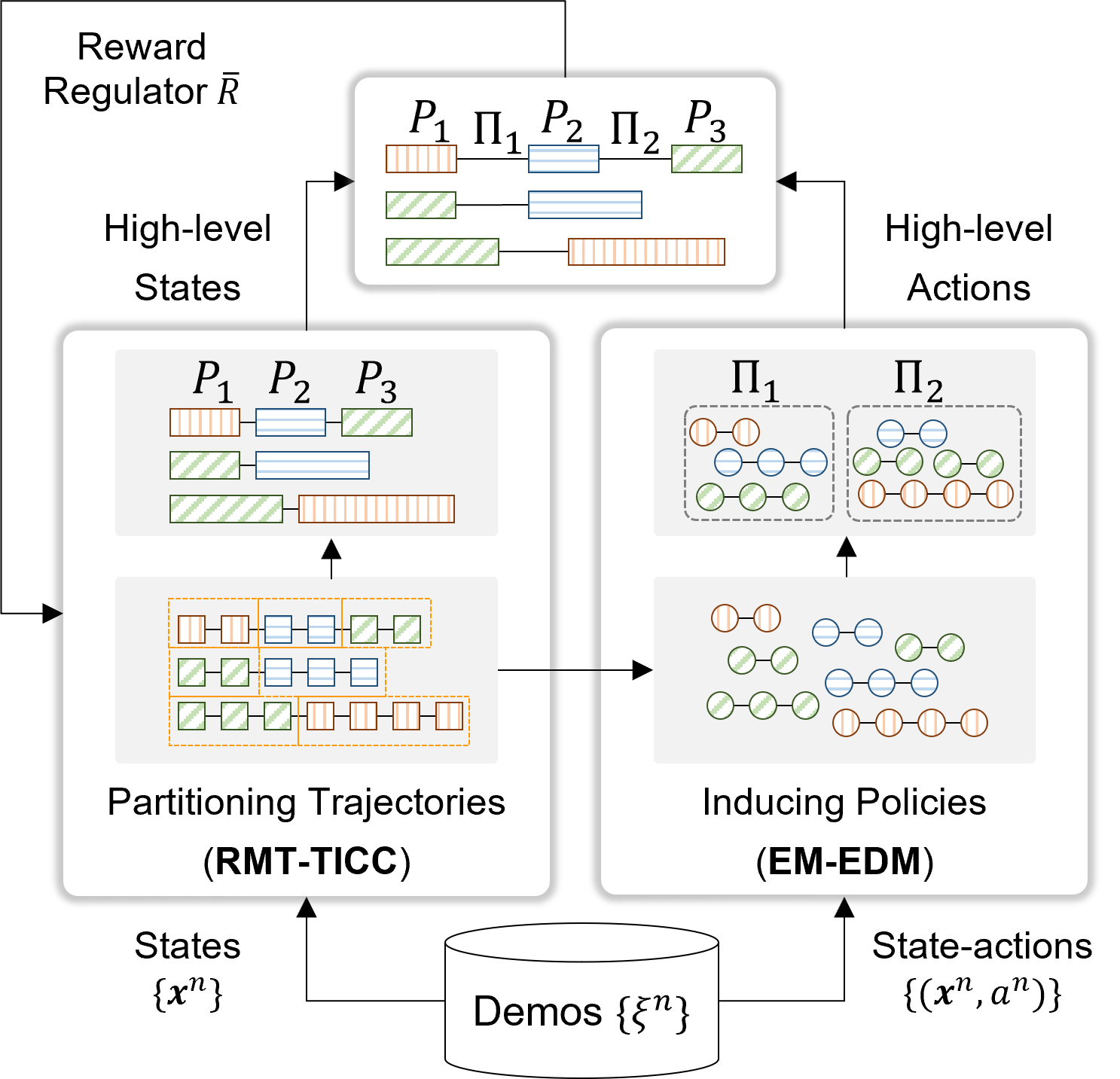}
    \caption{Overview of THEMES Framework.}
    \label{fig:overview}
\end{figure}


The input of THEMES contains $N$ demonstrated trajectories $\{\xi^{n}\} = \{(\mathbf{x}_t^n, a_{t}^{n})|t=1, ..., T^{n}; n = 1,..., N\}$ provided by experts, where $\mathbf{x}_t^n\in \mathbb{R}^m$ is the $t$-th multivariate state with $m$ features, and $a_{t}^{n}$ is the corresponding action in the $n$-th trajectory with the length of $T^{n}$. In this work, we assume that $\{\xi^{n}\}$ are carried out with \emph{multiple} policies that are driven by reward functions \emph{evolving} over time. To induce effective policies, THEMES would iteratively partition $\{\xi^{n}\}$ into sub-trajectories based on $\{\mathbf{x}^{n}\}$ and then cluster the sub-trajectories and induces policies for each cluster. Taking our EHRs as an example, THEMES will partition patients' sepsis progressive stages and induce clinicians' different treatment policies over different stages.

\begin{algorithm}
    \caption{\textbf{THEMES}}
    \label{alg:themes}
    \renewcommand{\thealgorithm}{}
    \begin{algorithmic}[1]
           \STATE \textit{Input}: Demonstrated trajectories $\{\xi^{n}\}$ \\ 
           \STATE \textit{Initial}: Reward regulator $\overline{R}=\mathbf{1}$
            \WHILE{$iter < \textit{Thres}$}
                \STATE Conduct \textbf{RMT-TICC} to learn
                \emph{high-level states} $\{P_{k}\}$
                \STATE Conduct \textbf{EM-EDM} to learn \emph{high-level actions} $\{\Pi_{g}\}$
                \STATE Learn a reward regulator $\overline{R}$ from $\{P_{k}\}$ and $\{\Pi_{g}\}$
            \ENDWHILE
            \STATE \textit{Output}: Sub-trajectory clusters with respective policies 
   \end{algorithmic}
\end{algorithm}

Figure \ref{fig:overview} shows the architecture of THEMES, which has a \emph{hierarchical} structure with two main components, \textit{i.e.}, RMT-TICC and EM-EDM, as detailed in Algorithm \ref{alg:themes}. 
Taking states $\{\mathbf{x}^{n}\}$ as input, the \textbf{RMT-TICC} will partition the trajectories, and the sub-trajectories are clustered so that each cluster shares the same time-invariant patterns over the states, which will be considered as a \emph{high-level state}. 
Then regarding the state-action pairs $\{(\mathbf{x}^{n}, a^{n})\}$ over the partitioned sub-trajectories, the \textbf{EM-EDM} will cluster and induce their policies, such that each cluster will share the consistent decision-making patterns, which will be considered as a \emph{high-level action}. 
Herein, we denote the high-level states as $\{P_{k}|k=1, ..., K\}$ and high-level actions as $\{\Pi_{g}|g=1, ..., G\}$, with $K$ and $G$ being the number of sub-trajectory clusters learned by RMT-TICC and EM-EDM, respectively. 
Then taking the high-level state-action pairs as input, a \emph{high-level reward regulator} $\overline{R}$ will be learned and fed back to refine the RMT-TICC. The whole procedure will be conducted interactively until convergence. Finally, THEMES will get fine-grained sub-trajectories clusters with their respective policies.

\subsection{RMT-TICC -- Partitioning Trajectories}

\subsubsection{Preliminaries} 

Given the \emph{states}  $\{\mathbf{x}_{t}^{n}|t=1, ..., T^{n}\}$ in $\{\xi^{n}\}$, our goal is to simultaneously \emph{partition and cluster} the sub-trajectories based on their latent time-invariant patterns, by learning a mapping from each state $\mathbf{x}_t^n$ to a certain cluster $\{k|k=1, ...,K\}$. 
Considering the interdependence among neighboring states, instead of treating each state independently, we explore the patterns within a sliding window $\omega \ll T^n$. Particularly, to determine which cluster $k$ the state $\mathbf{x}_t^n$ belongs to, the preceding states within $\omega$, \textit{i.e.}, $\mathbf{X}_t^n=\{\mathbf{x}_{t-\omega+1}^{n}, ...,\mathbf{x}_{t}^{n}\}$, will be considered. 
The $\mathbf{X}_t^n$ is represented as a $m\omega$-dimension random variable (obtained by concatenating the $m$-dim states in $\mathbf{X}_t^n$) and will be fit into $K$ Gaussian distributions, with the $k$-th fitted distribution corresponding to the $k$-th cluster.

To characterize sub-trajectory cluster distributions, Hallac \textit{et al.} proposed a Toeplitz inverse covariance-based clustering to learn the mean and inverse covariance matrix for each cluster \cite{hallac2017toeplitz}.  
Specifically, determining the mean vectors $\{\mathbf{\mu}_{k}| k=1,...,K\}$ is equivalent to matching each state to an optimal cluster, which leads to the clustering assignments $\mathbf{P}=\{P_k | k=1,...,K\}$, with $P_k \subset \{1, ..., T\}$ denoting the indices of states (sliding windows) belonging to cluster $k$. Meanwhile, the optimal inverse covariance matrices $\mathbf{\Theta} = \{\mathbf{\Theta}_k | k=1, ..., K\}$ are estimated to characterize the time-invariant structural patterns for each cluster. $\mathbf{\Theta}_{k}$ is constrained to be block-wise Toeplitz, composed of $\omega$ sub-blocks $A^{(i)}\in \mathbb{R}^{m\times m}$, $ i \in [0, \omega-1]$. The sub-block $A^{(i)}$ represents partial correlations among $m$ features between timestamp $t$ and $t+i$. For instance, the $(p,q)$-th entry in $A^{(i)}$ indicates the partial correlation between the $p$-th feature at $t$ and the $q$-th feature at $t+i$, where $p,q \in \{ 1, ..., m \}$. Note that the inverse covariance matrix is used rather than the covariance matrix because it models conditional dependencies and can easily introduce a graph structure, which can substantially decrease the number of parameters to avoid overfitting.

Based on the above theories, Yang \textit{et al.} proposed a Multi-series Time-aware Toeplitz Inverse Covariance-based Clustering (MT-TICC) approach \cite{yang2021multi}. The power of MT-TICC mainly exist in two aspects: 1) it takes \emph{multi-series} as input. Comparing to single time-series input, it can more accurately estimate the mean and inverse covariance matrix for each cluster based on the \textit{observed states} across different trajectories; and 2) it incorporates the \textit{time-awareness} to handle the irregular intervals, by employing a decay function to constrain the consistency within each cluster over time. These two properties are highly desired when modeling the real-world data such as EHRs. Note that in MT-TICC, the patterns for each cluster are learned from the \textit{observed states}. Since  \emph{decision-making patterns} conveyed by \emph{state-action pairs} usually have great impacts on the progression of trajectories \cite{komorowski2018artificial}, it is expected to take such patterns into account when learning the sub-trajectory clusters.

\subsubsection{RMT-TICC}

Since the interventions are driven by underlying \emph{reward functions}, to incorporate the \emph{decision-making patterns} when partitioning the trajectories, we introduce a \emph{reward regulator} to MT-TICC and propose a \emph{Reward-regulated} MT-TICC (RMT-TICC). Eq.\eqref{eq:4} shows the objective function of RMT-TICC: 

\vspace{-0.3cm}
\begin{align} 
\small 
    \begin{split}
    \argmin_{\mathbf{\Theta}, \mathbf{P}} \sum_{k=1}^{K}
        & \bigg[ 
        \sum_{n=1}^{N} \sum_{\mathbf{X}_{t}^{n} \in \mathbf{P}_{k}}
        \bigg(\overbrace{-\ell\ell(\mathbf{X}_{t}^{n}, \mathbf{\Theta}_{k})}^{\text{Log-likelihood}} + \\
        & \overbrace{c(\mathbf{X}_{t-1}^{n}, \mathbf{P}_{k}, \Delta T_{t}^{n}, \Delta r_{t}^{n})}^{\text{Reward-regulated Time-aware Consistency}} \bigg) 
         + \lambda \overbrace{||\mathbf{\Theta}_{k}||_{1}}^{\text{Sparsity}}\bigg] 
        \label{eq:4}
    \end{split}
\end{align}

\noindent $\bullet$ \textit{Log-likelihood term} measures the probability that $\mathbf{X}_{t}^{n}$ follows the $\mathbf{\Theta}_{k}$ and belongs to the cluster $k$. Assuming $\mathbf{X}_{t}^{n} \sim N(\mathbf{\mu}_{k}, \mathbf{\Theta}_{k}^{-1})$, $\ell\ell(\mathbf{X}_{t}^{n}, \mathbf{\Theta}_{k})$ is defined as:  
\vspace{-0.3cm}
\begin{align} 
\small
    \begin{split}
    \ell\ell(\mathbf{X}_{t}^{n}, \mathbf{\Theta}_{k}) = 
        & -\frac{1}{2}(\mathbf{X}_{t}^{n} - \mathbf{\mu}_{k})^{T} \mathbf{\Theta}_{k} (\mathbf{X}_{t}^{n} - \mathbf{\mu}_{k}) \\
        & + \frac{1}{2} \log |\mathbf{\Theta}_{k}| - \frac{m}{2}\log(2\pi)
        \label{eq:2}
    \end{split}
\end{align}

\noindent $\bullet$ \textit{Reward-regulated Time-aware Consistency term} differentiates our RMT-TICC from the orginal MT-TICC. It encourages consecutive states $\{\mathbf{X}_{t-1}, \mathbf{X}_{t}\}$ to be assigned into the same cluster while considering both the time intervals and the corresponding rewards. By minimizing Eq.\eqref{eq:3}, the neighbored states belonging to different clusters will be penalized. 
\begin{align} 
\small
    \begin{split}
    c(\mathbf{X}_{t-1}^{n}, \mathbf{P}_{k}, \Delta T^{n}_{t}, \Delta r^{n}_{t}) = \frac{\beta \mathds{1}\{t-1 \notin P_k\}}{\Phi(\Delta r^{n}_{t}, log(e+\Delta T^{n}_{t}))} 
        \label{eq:3}
    \end{split}
\end{align}

\noindent $\beta$ is a weight parameter. $\mathds{1}\{t-1 \notin P_k\}$ is an indicator function, with the value of 1 if $\mathbf{X}_{t-1}^{n}$ does not belong to the same cluster with $\mathbf{X}_{t}^{n}$; otherwise its value is 0. $1/log(e+\Delta T^{n}_{t})$ is a decay function, which can adaptively relax the penalization over the consistency constraint when the interval $\Delta T^{n}_{t}$ between consecutive states becomes larger \cite{baytas2017patient}.

To incorporate the decision-making patterns to refine the sub-trajectory partitioning, we employ a \emph{hierarchical} structure as shown in Figure \ref{fig:overview}. Instead of directly learning the rewards from observed state-action pairs, we employ such \emph{hierarchical} structure for two reasons: 1) the decision-making patterns across different sub-trajectories are of different importance, while directly learning from state-action pairs will treat them equally; 2) learning from state-action pairs in a flatten structure cannot fully capture the transitional patterns across different policies.

Given the high-level states $\{P_{k}\}$ and actions $\{\Pi_{g}\}$, a high-level reward function $\overline{R}$ will be learned accordingly. It is initialized as $\overline{R} = \mathbf{1}$. In each iteration, given the high-level state-action pairs, we employ a maximum likelihood inverse reinforcement learning \cite{babes2011apprenticeship} to update the $\overline{R}$, considering its efficiency in inferring reward functions \cite{yang2020student}. Based on $\overline{R}$, the reward $\{r_{t}^{n}|t=1,..., T^{n}\}$ for each state-action pair $(\mathbf{x}_{t}^{n}, a_{t}^{n})$ will be further calculated as: 

\vspace{-0.3cm}
\begin{align} 
\small
    \begin{split}
    r_{t}^{n} = \frac{1}{G} \sum_{g=1}^{G} \sum_{k=1}^{K} \mathds{1}\{t \in P_k\} \Pi_g(\mathbf{x}_{t}^{n}, a_{t}^{n}) \overline{R}(P_k, \Pi_g)
    \label{eq:2.1.4}
    \end{split}
\end{align}

\noindent Herein, 
$\mathds{1}\{t \in P_k\}$ has the value of 1 if $\mathbf{x}_{t}^{n}$ belongs to the sub-trajectory cluster $k$; otherwise its value is 0. $\Pi_{g}(\mathbf{x}_{t}^{n}, a_{t}^{n})$ denotes the probability of taking $a_{t}^{n}$ at $\mathbf{x}_{t}^{n}$ with the $g$-th policy. 
Based on the learned reward $r_{t}^{n}$, we further calculate the $\Delta r^{n}_{t}$ for consecutive state-action pairs to regulate the consistency constraint. 
To balance the effects of time-awareness patterns, \textit{i.e.}, $log(e+\Delta T^{n}_{t})$, and decision-making patterns, \textit{i.e.}, $\Delta r^{n}_{t}$, we employ a bivariate Gaussian distribution $\Phi$ to formulate their interactional regularizations. 


\noindent $\bullet$ \textit{Sparsity term} controls the sparseness based on a \textit{$l_{1}$}-norm, \textit{i.e.}, $\lambda || \mathbf{\Theta}_{k}||_{1}$, with $\lambda$ being a coefficient. It can select the most significant variables to represent the time-invariant structural patterns to effectively prevent overfitting.

To solve the objective function Eq.\eqref{eq:4}, we employ the EM to learn the cluster assignments $\mathbf{P}$ and the structural patterns $\mathbf{\Theta}$ iteratively until convergence. Specifically, \textit{in E-step}, by fixing $\mathbf{\Theta}$ to learn $\mathbf{P}$, Eq.\eqref{eq:4} degenerates into a form with only the log-likelihood term and the consistency term. It can be solved by dynamic programming to find a minimum cost \textit{Viterbi} path~\cite{viterbi1967error}; 
\textit{In M-step}, by fixing $\mathbf{P}$ to learn the $\mathbf{\Theta}$, Eq.\eqref{eq:4} degenerates into a form with only the log-likelihood term and the sparsity term. It can be formulated as a typical graphical lasso problem \cite{friedman2008sparse} with a Toeplitz constraint over $\mathbf{\Theta}$ and be solved by an alternating direction method of multipliers \cite{boyd2011distributed}.

\subsection{EM-EDM -- Inducing Policies}

\subsubsection{Preliminaries}

As a strictly \emph{offline} AL, energy-based distribution matching (EDM) \cite{jarrett2020strictly} can learn the policy merely based on the experts' demonstrations, not requiring any knowledge of model transitions or off-policy evaluations. It assumes that the demonstrations $\{\xi^{n}\}$ are carried out with a policy $\Pi^{\theta}$ parameterized by $\theta$, driven by a \emph{single} reward function. 

For simplicity, hereinafter, we denote the state-action pair as $(\mathbf{x}, a)$ by omitting their indexes when it does not cause ambiguity. 
The occupancy measures for the demonstrations and for the learned policy are denoted as $\rho_{\xi}$ and $\rho_{\Pi^{\theta}}$, respectively. 
The probability density for each state-action pair can be measured as:  $\rho_{\Pi^{\theta}}(\mathbf{x},a) =  \mathbb{E}_{\Pi^{\theta}}[\sum_{t=0}^{\infty} \gamma^{t} \mathds{1}{\{\mathbf{x}_{t}=\mathbf{x}, a_{t}=a\}}]$, where $\gamma$ is a discount factor, then the probability density for each state can be measured by: $\rho_{\Pi^{\theta}}(\mathbf{x}) = \sum_{a} \rho_{\Pi^{\theta}}(\mathbf{x},a)$. To induce the policy $\Pi^{\theta}$, our goal is minimizing the KL divergence between $\rho_{\xi}$ and $\rho_{\Pi^{\theta}}$: 

\vspace{-0.2cm}
\begin{align} 
\small
    \begin{split}
    \argmin_{\theta} D_{KL}(\rho_{\xi}||\rho_{\Pi^{\theta}}) = \argmin_{\theta} -\mathbb{E}_{\mathbf{x},a \sim \rho_{\xi}} \log \rho_{\Pi^{\theta}}(\mathbf{x},a)
    \end{split}
\end{align}

Since $\Pi^{\theta}(a|\mathbf{x}) = \rho_{\Pi^{\theta}}(\mathbf{x},a)/\rho_{\Pi^{\theta}}(\mathbf{x})$, we can formulate the objective function as: 

\vspace{-0.2cm}
\begin{equation}
\small
\label{eq:6}
    \argmin_{\theta} - \mathbb{E}_{\mathbf{x} \sim \rho_{\xi}} \log \rho_{\Pi^{\theta}}(\mathbf{x}) - \mathbb{E}_{\mathbf{x},a \sim \rho_{\xi}} \log \Pi^{\theta}(a|\mathbf{x}) 
\end{equation} 

\noindent When there is no access to roll out the policy $\Pi^{\theta}$ in an \emph{online} manner, $\rho_{\Pi^{\theta}}(\mathbf{x})$ in the first term of Eq.\eqref{eq:6} would be difficult to estimated. EDM can handle this issue utilizing an energy-based model \cite{grathwohl2019your}.

According to energy-based model, the probability density $\rho_{\Pi^{\theta}}(\mathbf{x}) \propto e^{-E(\mathbf{x})}$, with $E(\mathbf{x})$ being an energy function. 
Then the occupancy measure for state-action pairs can be represented as:  $\rho_{\Pi^{\theta}}(\mathbf{x},a) = e^{f_{\Pi^{\theta}}(\mathbf{x}) [a]} / Z_{\Pi^{\theta}}$, and the occupancy measure for states can be obtained by marginalizing out the actions: $\rho_{\Pi^{\theta}}(\mathbf{x}) = \sum_{a} e^{f_{\Pi^{\theta}}(\mathbf{x}) [a]} / Z_{\Pi^{\theta}}$. Herein, $Z_{\Pi^{\theta}}$ is a partition function, and $f_{\Pi^{\theta}}: \mathbb{R}^{|X|} \rightarrow \mathbb{R}^{\mathbb{|A|}}$ is a parametric function that mapping each state to $\mathbb{A}$ real-valued numbers. 


The parameterization of $\Pi^{\theta}$ implicitly defines an energy-based model over the states distribution, where the energy function can be defined as: $E_{\Pi^{\theta}}(\mathbf{x}) = -\log \sum_{a} e^{f_{\Pi\theta}(\mathbf{x})[a]}$. 
Under the scope of energy-based model, the first term in Eq.\eqref{eq:6} can be reformulated as an occupancy loss: 

\vspace{-0.2cm}
\begin{equation}
\label{eq:8}
    \mathcal{L}_{\rho}(\theta) = \mathbb{E}_{\mathbf{x} \sim \rho_{\xi}} E_{\Pi^{\theta}} (\mathbf{x}) - \mathbb{E}_{\mathbf{x} \sim \rho_{\Pi^{\theta}}} E_{\Pi^{\theta}} (\mathbf{x}) 
\end{equation}

\noindent where $\nabla_\theta \mathcal{L}_\rho (\theta) = - \mathbb{E}_{\mathbf{x} \sim \rho_{\xi}} \nabla_\theta \log \rho_{\Pi^{\theta}}(\mathbf{x})$ can be solved by existing optimizers, \textit{e.g.}, stochastic gradient Langevin dynamics \cite{welling2011bayesian}. Therefore, by substituting the first term in Eq.\eqref{eq:6} as Eq.\eqref{eq:8} via energy-based model, we can derive a \emph{surrogate objective function} to get the optimal solution without the need of \emph{online} rolling out the policy. 


\subsubsection{EM-EDM}

To deal with \emph{multiple} reward functions varying across the demonstrations, Babes-Vroman \textit{et al.} proposed an EM-based inverse reinforcement learning \cite{babes2011apprenticeship}, by iteratively clustering the demonstrations in \textit{E-step} and inducing policies for each cluster by IRL in \textit{M-step}. Specifically, in the \textit{M-step}, they explored several IRL methods based on the \emph{discrete} states, which is not scalable for large continuous state spaces, \textit{e.g.}, EHRs. 
Enlightened by this EM framework and the success of EDM, we propose an EM-EDM.

Taking sub-trajectories $\{\hat{\xi}^{\hat{n}} | \hat{n}=1,...,\hat{N}\}$ learned by RMT-TICC as input, with $\hat{N}$ being the number of sub-trajectories, EM-EDM aims to cluster these sub-trajectories and learn the cluster-wise policies $\{\Pi_g|g=1,...G\}$, where $G$ is the number of clusters. $\nu_{g}$ and $\theta_{g}$, denote the prior probability and the policy parameter for each cluster, and both are randomly initialized.  
The objective function of EM-EDM is to maximize the log-likelihood defined in Eq.\eqref{eq:10}:


\begin{equation}
\small
\label{eq:10}
    \argmax_{\theta_g} \underset{g=1}{\overset{G}{\sum}} \underset{\hat{n}=1}{\overset{\hat{N}}{\sum}} log(u_{\hat{n}g})
\end{equation}

\noindent where $u_{\hat{n}g}$ denotes the probability that trajectory $\hat{\xi}^{\hat{n}}$ follows the policy of the $g$-th cluster. It is defined in Eq.\eqref{eq:5}, with $U$ being a normalization factor.

\begin{equation}
\small
\label{eq:5}
     u_{\hat{n}g} = Pr(\hat{\xi}^{\hat{n}}|\theta_{g}) = \underset{(\mathbf{x},a) \in \hat{\xi}^{\hat{n}}}{\prod} \frac{\Pi_{\theta_{g}}(\mathbf{x},a)\nu_{g}}{U},
\end{equation}

During the EM process, in the \textit{E-step}, the probability that trajectory $\hat{\xi}^{\hat{n}}$ belonging to cluster $g$ is calculated by Eq.\eqref{eq:5}. 
Then in the \textit{M-step}, the prior probabilities are updated via $\nu_{g} = \sum_{\hat{n}} u_{\hat{n}g}/\hat{N}$, and the policy parameters $\theta_{g}$ is learned by EDM. 
The \textit{E-step} and \textit{M-step} are iteratively executed until converged. Finally, the output of EM-EDM is the clustered trajectories with their respective policies.


\section{Experiments}

To validate the effectiveness of the THEMES framework, we applied it to a healthcare application with electronic healthcare records (EHRs), which are comprehensive longitudinal collections of patients' data that play a critical role in modeling disease progression to facilitate clinical decision-making. Our EHRs were collected by the Christiana Care Health System in the U.S. over a period of two and a half years.

\subsection{Data Preprocessing}
\label{sec:preprocessing}
We identified 52,919 patients' visits (4,224,567 timestamps) with suspected infection as the sepsis-related study cohort. Note that the rules employed for identifying suspected infection and tagging septic shock were provided by two leading clinicians with extensive sepsis experience. The selected cohort was preprocessed by the following steps: 

\noindent $\bullet$ \textit{Feature selection}: 14 sepsis-related features were selected as suggested by clinicians, including: 
1) \textit{Vital signs}: systolic blood pressure, mean arterial pressure, respiratory rate, oxygen saturation, heart rate, temperature, the fraction of inspired oxygen; and  
2) \textit{Lab results}: white blood cell, bilirubin, blood urea nitrogen, lactate, creatinine, platelet, neutrophils.  

\noindent $\bullet$ \textit{Missing data handling}: 
The timestamps in each visit were collected with irregular intervals, ranging from 0.94 seconds to 28.19 hours. Different features are measured with varying frequencies, which causes some features to be unavailable and missing from certain timestamps. On average, the missing rate of our data is $\sim$80.37\%. Herein, we handled the absence of data by carrying forward, \textit{i.e.}, filling the missing entries as the last observation until the next observed value, with the remaining missing entries filled with the mean value. 

\noindent $\bullet$ \textit{Tagging the septic shock visits}: 
Identifying the septic shock visits is a challenging task. Though the diagnosis codes, \textit{e.g.}, ICD-9, are widely used for clinical labeling, solely relying on the codes can be problematic: they have proven to be limited in reliability since the coding practice is mainly used for administrative and billing purpose \cite{zhang2019attain}. 
Based on the Third International Consensus Definitions for Sepsis and Septic Shock \cite{singer2016third}, our domain experts identified septic shock when either of the following two conditions was met: 
1) Persistent hypertension through two consecutive readings ($\le$ 30 minutes apart), including systolic blood pressure(SBP) $<$ 90 mmHg, mean arterial pressure $<$ 65mmHg, and decrease in SBP $\geq$ 40 mmHg within an 8-hour period; or 2) Any vasopressor administration. By combing both ICD-9 codes and experts' rules, we identified 1,869 shock and 23,901 non-shock visits.

\subsection{Selecting Demonstrations}
In AL, it is usually assumed that the demonstrations are optimally or near optimally executed by experts \cite{abbeel2004apprenticeship}. Thus, the quality of the demonstrations matters to induce more accurate policies. To filter higher-quality demonstrations, we designed a strategy to evaluate the effectiveness of interventions by early prediction. Specifically, given the trajectories truncated \emph{before the first intervention}, we trained 100 effective long short-term memory (LSTM) classifiers \cite{baytas2017patient} for early predicting shock/non-shock using different training data and hyperparameter settings. The trajectories on which more than 80\% predictors reporting the same results were kept. Comparing the early prediction results vs. the outcomes at onset: if a patient converts from shock to non-shock, it indicates the interventions would be effective, then the trajectory will be considered with higher quality and will be taken as experts' demonstrations; Otherwise, the trajectories will be filtered out from the analysis. 

Following the above steps, we identified 195 trajectories as the experts' demonstrations. Based on these demonstrations, we modeled the states and actions as follows: 

\noindent $\bullet$ \textbf{States} are defined based on the 14 continuous features related to sepsis progression introduced in Sec.\ref{sec:preprocessing}. Except for the original features, for each feature per timestamp, we also calculated the max and min values that happened in the past 1 hour. In total, the states are formulated as 42-dim vectors. 

\noindent $\bullet$ \textbf{Actions} are binary indicators characterizing whether antibiotics (\textit {e.g.}, clindamycin, daptomycin) are scheduled by clinicians or not. As suggested by Gauer \cite{gauer2013early}, antibiotic therapy plays an important role in improving the clinical outcomes for sepsis patients.

\subsection{Experimental Settings}

\renewcommand{\arraystretch}{1.2}
\setlength\tabcolsep{3pt}
\begin{table*}
\footnotesize
    \centering
    \setlength\tabcolsep{5pt}
    \begin{tabular}{c|ccccccc|} \hline
        \textbf{Methods} & \textbf{Acc} & \textbf{Rec} & \textbf{Prec} & \textbf{F1} & \textbf{AUC} & \textbf{Jaccard} \\
        \hline
        GP\&DQN \cite{azizsoltani2019unobserved} & .588(.042) & .342(.033) & .246(.045) & .286(.045) & .552(.037) & .167(.028) \\ 
        LSTM \cite{duan2017one} & .553(.043) & .441(.080) & .244(.034) & .312(.044) & .537(.033) & .186(.031)  \\ 
        EDM \cite{jarrett2020strictly} & .740(.022) & .760(.058) & .462(.034) & .572(.014) & .817(.014) & .400(.014) \\ 
        HIRL \cite{krishnan2016hirl} & .788(.014) & .733(.016) & .520(.015) & .608(.012) & .844(.014) & .437(.012) \\
        MIL \cite{hausman2017multi} & \textbf{.803(.013)} & \textbf{.762(.042)} & \textbf{.538(.032)} & \textbf{.631(.012)} & \textbf{.872(.012)} & \textbf{.476(.014)} \\  \hline
        
        EM-EDM & .748(.018) & .748(.036) & .473(.033) & .578(.020) & .807(.012) & .407(.019) \\ 
        MT-TICC\&EDM & .836(.019) & .659(.045) & .650(.037) & .646(.014) & .869(.011) & .481(.014) \\ 
        THEMES\_0 & \textbf{.865(.013)} & \textbf{.781(.058)} & \textbf{.678(.038)} & \textbf{.723(.012)} & \textbf{.904(.010)} & \textbf{.566(.015)} \\ \hline 
        
        \textbf{THEMES} & \textbf{.867(.012)}* & \textbf{.830(.010)}* & \textbf{.685(.020)}* & \textbf{.751(.010)}* & \textbf{.905(.008)}* & \textbf{.600(.013)}* \\ \hline 
    \end{tabular}
    \caption{\centering Comparing THEMES to five baselines and three ablations. The best methods are in bold, and the overall best is highlighted with *. }
    \label{tab:themes}
\end{table*}

To evaluate the effectiveness of THEMES, we compared it against five baselines as well as three ablation approaches. 

\noindent The five baselines include: 

\noindent $\bullet$ \textbf{GP\&DQN} \cite{azizsoltani2019unobserved}, which is not an AL approach but rather a deep reinforcement learning approach. It employs a Gaussian process (GP) to infer the rewards from the patients' final outcomes, \textit{i.e.}, shock/non-shock, then applies the Deep Q-Network (DQN) \cite{fan2020theoretical} to learn a policy. This approach is selected because much prior work on sepsis treatments leverages offline DQN to induce policies. 

\noindent $\bullet$ \textbf{LSTM} \cite{duan2017one}, which is a \emph{behavior cloning offline AL} method that leverages LSTM to learn policy as a mapping from states to actions.

\noindent $\bullet$ \textbf{EDM} \cite{jarrett2020strictly}, the state-of-the-art offline AL. It has been demonstrated that EDM can outperform competitive cutting-edge AL methods with a \emph{single} reward function, thus we will not repetitively conduct those comparisons here. 

\noindent $\bullet$ \textbf{HIRL} \cite{krishnan2016hirl} is adapted from a hierarchical inverse reinforcement learning (HIRL), which learns the sub-trajectory clusters by Gaussian mixture model \cite{reynolds2009gaussian}. Instead of following the original HIRL work to learn cluster-wise policies by maximum entropy IRL \cite{ziebart2008maximum}, here we applied the more powerful EDM.


\noindent $\bullet$ \textbf{MIL} \cite{hausman2017multi}, which is a multi-modal imitation learning (MIL), by modeling each sub-trajectory cluster as a modality. It employs the infoGAN \cite{chen2016infogan} to learn the sub-trajectory clusters with their respective policies.


\noindent The three ablation methods of THEMES include: 

\noindent $\bullet$ \textbf{EM-EDM}, which assumes the demonstrations follow \emph{multiple} reward functions varying across trajectories (while remaining the same within each trajectory). 

\noindent $\bullet$ \textbf{MT-TICC\&EDM}, which learns sub-trajectory clusters by MT-TICC, then induces the policy for each cluster by EDM. 

\noindent $\bullet$ \textbf{THEMES\_0}, which can be regarded as a simplified version of THEMES without iterative refinement by the reward regulator, \textit{i.e.}, with 0 high-level iterations. It learns sub-trajectory partitions by MT-TICC and then induces policies across the sub-trajectories by EM-EDM.

Each of the above methods was repeated 10 times by randomly splitting 80\% data for training and 20\% for testing. We employed the metrics of Accuracy (Acc), Recall (Rec), Precision (Prec), F1-score (F1), AUC, and Jaccard score for evaluation. All the model parameters were determined by 5-fold cross-validation. Particularly, in THEMES, for RMT-TICC, based on Bayesian information criteria (BIC) \cite{watanabe2013widely}, we determined the cluster number $K$ as 11, the window size $\omega$ as 2, and the sparsity and consistency coefficients $\lambda$ and $\beta$ as 1e-5 and 4, respectively. For EM-EDM, the optimal cluster number was determined heuristically as 3, by iteratively implementing the EM, until empty clusters are generated, or the log-likelihood of the clustering results varied smaller than a pre-defined threshold. The iterations for the overall THEMES framework had a threshold of 10, based on our observation that the clustering likelihood for both MT-TICC and EM-EDM get converged within 10 iterations. For fair comparisons, the optimal parameters in other baselines and ablations were determined by cross-validation as well.

\subsection{Results}

\begin{figure}
     \centering
     \begin{subfigure}
         \centering
         \includegraphics[width=8cm]{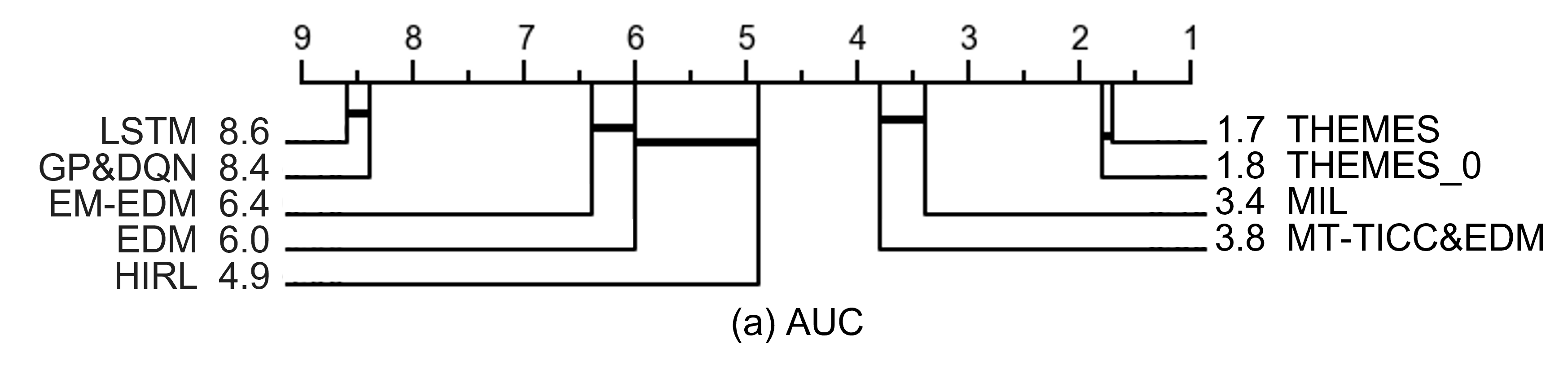}
     \end{subfigure}
     \hfill
     \begin{subfigure}
         \centering
         \includegraphics[width=8cm]{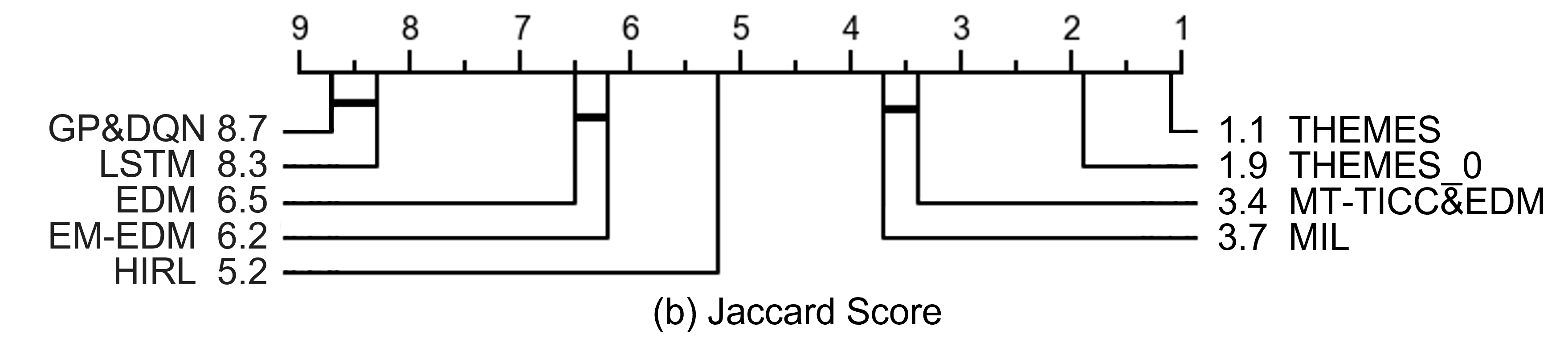}
     \end{subfigure}
    \vspace{-0.5cm}
    \caption{Critical difference diagram with Wilcoxon signed-rank test over (a) AUC and (b) Jaccard Score.}
    \label{fig:cd_graph}
\end{figure}

Our results are reported in Table \ref{tab:themes}. The best results among baselines and ablations are in bold, and the overall best results are highlighted with *. To better evaluate the significance, we plot the critical difference diagrams with Wilcoxon signed-rank tests over AUC and Jaccard score, as shown in Figure \ref{fig:cd_graph}, where the unconnected models indicate pairwise significance, with a confidence level of 0.05. The results show THEMES performs the best in terms of all evaluation metrics.

\noindent $\bullet$ \textit{THEMES vs. Five Baselines}: Among the five baselines, MIL performs the best. While as shown in Table \ref{tab:themes}, THEMES outperforms all five baselines across all evaluation metrics and the differences are significant according to  Figure \ref{fig:cd_graph}. It indicates the effectiveness of \emph{time-awareness} in THEMES, which cannot be captured by any of the five baselines. 
Meanwhile, THEMES, MIL, and HIRL outperform the other three baselines that assuming a \emph{single} reward function, which indicates that \emph{multiple evolving} reward functions would be a better way for modeling complex human-centric tasks, such as healthcare. Furthermore, all AL methods in general outperform the non-AL method, \textit{i.e.}, GP\&DQN, in which the policy relies on the accuracy of the GP-inferred rewards, and it requires a large amount of training data for DQN to induce an effective policy. Besides, among all AL methods, LSTM performs the worst, which is probably because the direct behavior cloning by LSTM cannot take the state distribution into account when learning the policy.

\vspace{0.03in}

\noindent $\bullet$ \textit{THEMES vs. Ablations}: Among the three ablation methods, THEMES\_0 performs the best, and THEMES outperforms THEMES\_0, which indicates the effectiveness of the hierarchically learned reward regulator in THEMES to incorporate \emph{decision-making patterns} when partitioning the trajectories. In addition, both THEMES and THEMES\_0 outperform MT-TICC\&EDM with significance, which reflects the power of EM-EDM in inducing policies over the sub-trajectories. Moreover, the significant improvements when comparing THEMES and THEMES\_0 against EM-EDM indicate the power of automatically partitioning trajectories by RMT-TICC or MT-TICC, which can effectively capture the different progressive patterns across sub-trajectory clusters. 

\vspace{0.03in}

Finally, when comparing EM-EDM to EDM, they do not show significant differences. Meanwhile, they perform not as well as THEMES and the other two THEMES ablation models. It indicates that the assumption of \emph{multiple} reward functions varying across trajectories (while remaining the same within each trajectory) in EM-EDM cannot fully model the \emph{evolving} decision-making patterns in EHRs.

\section{Conclusions and Discussions}

In this paper, we propose THEMES framework to induce policies from the experts' demonstrations in an \emph{offline} manner when there are \emph{multiple} reward functions \emph{evolving} over time. The effectiveness of the proposed framework is evaluated via a healthcare application to induce policies from clinicians when treating an extremely challenging disease, \textit{i.e.}, sepsis. The results demonstrate that THEMES can outperform competitive baselines in policy induction. Especially, empowered by time-awareness and decision-making patterns encoded in a hierarchically learned reward regulator, more fine-grained sub-trajectory partitions can be learned, based on which more accurate policies can be further induced. 

The major limitations of THEMES include: 1) When modeling time-invariant patterns to partition sub-trajectories in RMT-TICC, the method employs a fixed-length sliding window. This can be further improved with adaptive window size and attention networks. 2) The framework is developed for modeling discrete actions. Further explorations are needed to fit continuous actions. 3) Since there are no standard human-centric benchmarks with \emph{multiple} reward functions \emph{evolving} over time, we validated THEMES based on a complex EHRs dataset. More experiments will be conducted in future work with the emergence of other feasible real-world datasets.

\bibliographystyle{named}
\bibliography{ijcai23.bib}

\end{document}